\documentclass{article}

\usepackage{arxiv}
\usepackage[utf8]{inputenc} 
\usepackage[T1]{fontenc}    
\usepackage{hyperref}       
\usepackage{url}            
\usepackage{booktabs}       
\usepackage{amsfonts}       
\usepackage{nicefrac}       
\usepackage{microtype}      
\usepackage{lipsum}
\usepackage{graphicx}
\graphicspath{ {./images/} }

\usepackage{color}
\usepackage{xcolor}
\usepackage{graphicx}
\usepackage{booktabs}
\usepackage{multirow}
\usepackage{amsmath,amsfonts,bm}

\usepackage{amsmath,amsfonts,bm}
\DeclareMathAlphabet{\mathsfit}{\encodingdefault}{\sfdefault}{m}{sl}
\SetMathAlphabet{\mathsfit}{bold}{\encodingdefault}{\sfdefault}{bx}{n}
\newcommand{\tens}[1]{\bm{\mathsfit{#1}}}

\def\tE{{\tens{E}}}
\def\tF{{\tens{F}}}

\def\tP{{\tens{P}}}
\def\tQ{{\tens{Q}}}
\def\tR{{\tens{R}}}
\def\tS{{\tens{S}}}

\def\tW{{\tens{W}}}

\title{MGMapNet: Multi-Granularity Representation Learning for End-to-End Vectorized HD Map Construction}

\author{
 Jing Yang \footnotemark[1] ~   \footnotemark[3]\\
  	Tongji University\\
   \And
 Minyue Jiang \footnotemark[1]\\
  Baidu Inc.\\
  \And
 Sen Yang \footnotemark[1]\\
  Baidu Inc.\\
  \And
 Xiao Tan \\
  Baidu Inc.\\
  \And
 Yingying Li \\
  Baidu Inc.\\
  \And
  Errui Ding \\
  Baidu Inc.\\
  \And
     \\[3pt]
  \And
  Hanli Wang \footnotemark[2]\\
  Tongji University\\
  \And
  Jingdong Wang \footnotemark[2]\\
  Baidu Inc.\\
   }

\begin{document}
\maketitle

\renewcommand{\thefootnote}{\fnsymbol{footnote}} 
\footnotetext[1]{These authors contributed equally to this work.} 
\footnotetext[2]{Corresponding authors.} 
\footnotetext[3]{Work done during an internship at Baidu.} 
\begin{abstract}
The construction of Vectorized High-Definition (HD) map typically requires capturing both category and geometry information of map elements. Current state-of-the-art methods often adopt solely either point-level or instance-level representation, overlooking the strong intrinsic relationships between points and instances. In this work, we propose a simple yet efficient framework named MGMapNet (Multi-Granularity Map Network) to model map element with a multi-granularity representation, integrating both coarse-grained instance-level and fine-grained point-level queries. Specifically, these two granularities of queries are generated from the multi-scale bird's eye view (BEV) features using a proposed Multi-Granularity Aggregator. In this module, instance-level query aggregates features over the entire scope covered by an instance, and the point-level query aggregates features locally. Furthermore, a Point Instance Interaction module is designed to encourage information exchange between instance-level and point-level queries. Experimental results demonstrate that the proposed MGMapNet achieves state-of-the-art performance, surpassing MapTRv2 by 5.3 mAP on nuScenes and 4.4 mAP on Argoverse2 respectively.
\end{abstract}


\section{Introduction}
Perceiving and understanding road map elements are crucial for ensuring the safety in autonomous driving applications~\cite{xiao2020multimodal,xu2023drivegpt4,prakash2021multi}. High-Definition (HD) maps provide category and geometry information about road elements, enabling autonomous vehicles to maintain lane position, anticipate intersections, and plan optimal routes to mitigate potential risks. However, constructing HD map requires significant human effort for annotating and updating, which limits scalability over large areas. Recent research, such as~\cite{HDMapNet, liao2022maptr, maptrv2, ding2023pivotnet, streammapnet, hu2021fiery}, focuses on learning-based methods as alternatives to construct HD map from onboard sensors.
These methods can be mainly divided into two categories based on the representation in use: rasterized map based representation~\cite{HDMapNet,li2022bevformer,liu2023bevfusion,xiong2023neural} and vectorized map based representation~\cite{ding2023pivotnet,li2023lanesegnet,maptrv2}. 

Rasterized map based methods often require complex post-processing to meet the need of downstream modules, such as planning. Consequently, this process may result in suboptimal results that are not entirely end-to-end optimized. Therefore, there has been increasing attention paid to end-to-end map construction methods~\cite{shin2023instagram,qiao2023machmap,zhang2024online} using vectorized representations.

Vectorized map based methods commonly employ Bird's Eye View (BEV)~\cite{fadadu2022multi,chen2017multi,liang2019multi,you2019pseudo,liang2018deep} space for end-to-end perception, effectively integrating various sensor information such as surround-view cameras and Lidar. State-of-the-art (SOTA) methods typically adopt DETR-like architectures, comprising encoder and decoder components. The encoder initially extracts multi-sensor information into BEV representation, while the decoder subsequently decodes the category and geometry information of each road element through queries. These methods achieve an end-to-end vectorized representation of output map elements, eliminating the need for the complex post-processing steps involved in rasterized maps representation.

SOTA methods use either point-level queries or instance-level queries  to generate map elements. Point-level queries are good at describing the geometric position of road elements. For instance, in MapTR~\cite{liao2022maptr}, a permutation-equivalent point expression accurately represents the location information of map elements, ensuring stable training processes. MapTRv2~\cite{maptrv2} further enhances perception accuracy by incorporating one2many training strategy and decoupled self-attention operation. However, these methods may lack an overall description of map elements, leading to deficiencies in representing lane relationships. For example, MapTRv2 may miss lane lines in distant and merging scenarios, as the region illustrated in the purple ellipses of Fig.~\ref{fig:vis_compare}.

While instance-level queries excel at capturing the overall category information of a road element, they may struggle to accurately represent geometric details, especially for irregular or elongated map elements. For example, in StreamMapNet~\cite{streammapnet}, the Multi-Point Attention mechanism is proposed to capture the overall information of road elements, allowing for longer attention ranges while maintaining computational efficiency. However, this method may encounter difficulties in accurately perceiving the geometry of irregular or elongated elements, leading to local disturbances. The green boxes in Fig.~\ref{fig:vis_compare} highlight the issue of inaccurate point coordinates obtained from instance-level queries, where the map elements are detected though, their positional accuracy is compromised.

To simultaneously incorporate both fine-grained local positions and coarse-grained global classification information, we propose a simple framework called MGMapNet (Multi-Granularity Map Network), which represents map elements through multi-granularity queries. Within each decoder layer, point-level queries and instance-level queries are simultaneously computed by querying the multi-scale bird’s eye view (BEV) features using Multi-Granularity Aggregator. Subsequently, Point Instance Interaction, including point-to-point attention and point-to-instance attention, is designed to enhance the intrinsic relationships. Ultimately, point-granularity queries are utilized for localizing point coordinates, while instance-granularity queries are employed for determining the categories of map elements.

Our main contributions can be summarized as follows:
\begin{figure}[t]
    \centering
    \includegraphics[width=1\textwidth]{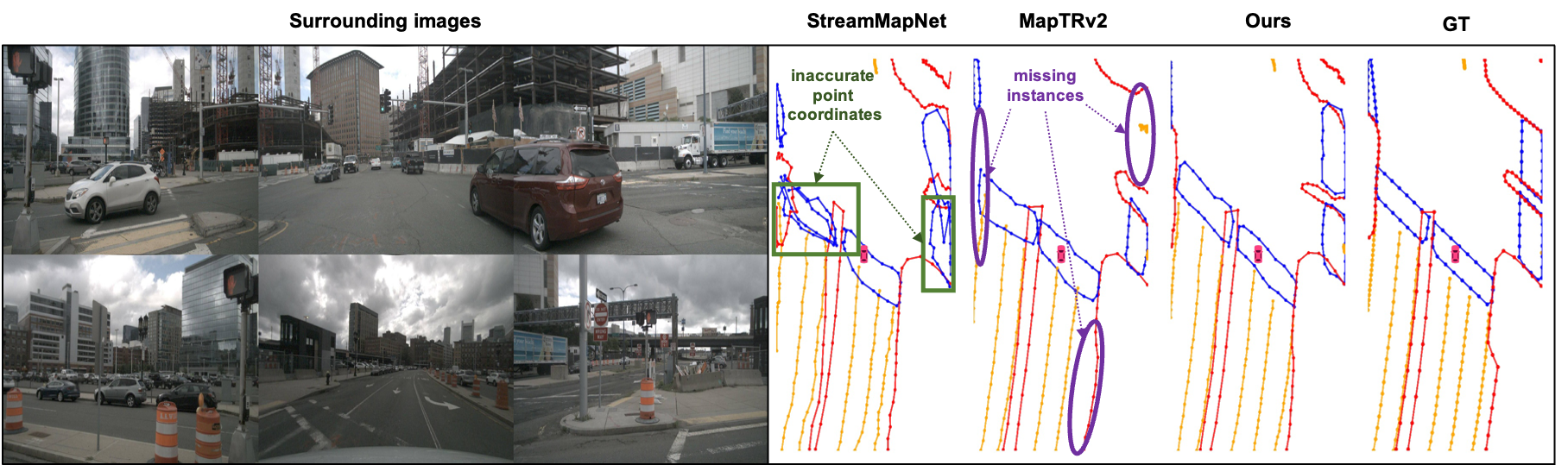}
    \caption{\textbf{Comparison of visualization results.} Visual comparison among StreamMapNet, MapTRv2, and MGMapNet, with vehicle-centric views featuring \textcolor{red}{red boundaries}, \textcolor{orange}{orange dividers}, and \textcolor{blue}{blue pedestrian crossings}. The \textcolor{green}{green boxes} denote the phenomenon of inaccurate point coordinates in instance-level queries, while the \textcolor{purple}{purple ellipses} indicate the phenomenon of missing instances in point-level queries. StreamMapNet employs MPA for single-frame results. Best viewd in color.}
    \label{fig:vis_compare}
\end{figure}
\begin{itemize}
\item We propose a robust multi-granularity representation, enabling the end-to-end construction of vectorized HD map by employing coarse-grained instance-level and fine-grained point-level queries in one framework.
\item The Multi-Granularity Aggregator, combined with Point Instance Interaction, facilitates an efficient interaction between point-level and instance-level queries, effectively exchanging category and geometry information.
\item We incorporated several strategy optimizations into the training, enabling our proposed MGMapNet to achieve state-of-the-art (SOTA) single-frame performance on both the nuScenes and Argoverse2 datasets.
\end{itemize}

\section{Related Work}
\label{gen_inst}

\paragraph{Online HD Map Construction.}
In recent years, researchers have increasingly utilized onboard sensors in autonomous driving to construct HD map. Previous work~\cite{huang2023anchor3dlane}\cite{chen2022persformer} has focused on projecting and lifting map elements detected on the Perspective View (PV) plane into 3D space for map reconstruction. With the aim of better integrating multiple sensors such as panoramic cameras and LiDAR, construction methods for online HD map are gradually transitioning to BEV representation. Currently, HD map construction can be broadly categorized into two types: rasterized map-based and vectorized map-based methods. Rasterized methods, such as HDMapNet~\cite{HDMapNet}, utilize BEV features for semantic segmentation, followed by a post-processing step to obtain vectorized map instances. Similarly, BEV-LaneDet~\cite{Wang_2023_CVPR} outputs confidence scores, embeddings for clustering, y-axis offsets, and average heights for each grid. While rasterized maps can provide detailed road information, the requirement for post-processing limits their applications. With the emergence of vectorized DETR-like~\cite{carion2020end} end-to-end methods, the need for post-processing is eliminated.
VectorMapNet~\cite{liu2023vectormapnet} is the first end-to-end map reconstruction model that utilizes transformers.
MapTR and MapTRv2~\cite{liao2022maptr,maptrv2} introduce a novel and unified modeling method for map elements, addressing ambiguity and ensuring stable learning processes. PivotNet~\cite{ding2023pivotnet} employs unified, pivot-based representations for map elements and is formulated as a direct set prediction paradigm. 

However, these methods often exclusively use either point-level queries or instance-level queries, missing out on the mutual advantages of both granularities. To address this limitation, this paper introduces a multi-granularity mechanism for representing map elements. This mechanism adaptively derives features at both fine-grained point granularity and coarse-grained instance granularity, thus preserving local details as well as global map information.

\paragraph{Lane Detection.}
Lane detection can be regarded as a subtask of high-definition map construction, focusing on the detection of lane elements within road scenes.
Current methods~\cite{li2019line,zheng2022clrnet,tabelini2021polylanenet} predominantly engage in lane detection from a single perspective view (PV) image, and the majority of lane detection datasets provide annotations only from a single perspective.
LaneATT~\cite{tabelini2021keep} proposes a novel anchor-based attention mechanism that aggregates global information. 
Unlike lane detection, vectorized HD map construction involves more complex map elements within the vehicle's perception range, including lane markings, curbs, and sidewalks.

\section{Method}
\label{headings}

\begin{figure}[t]
    \centering
    \includegraphics[width=1\textwidth]{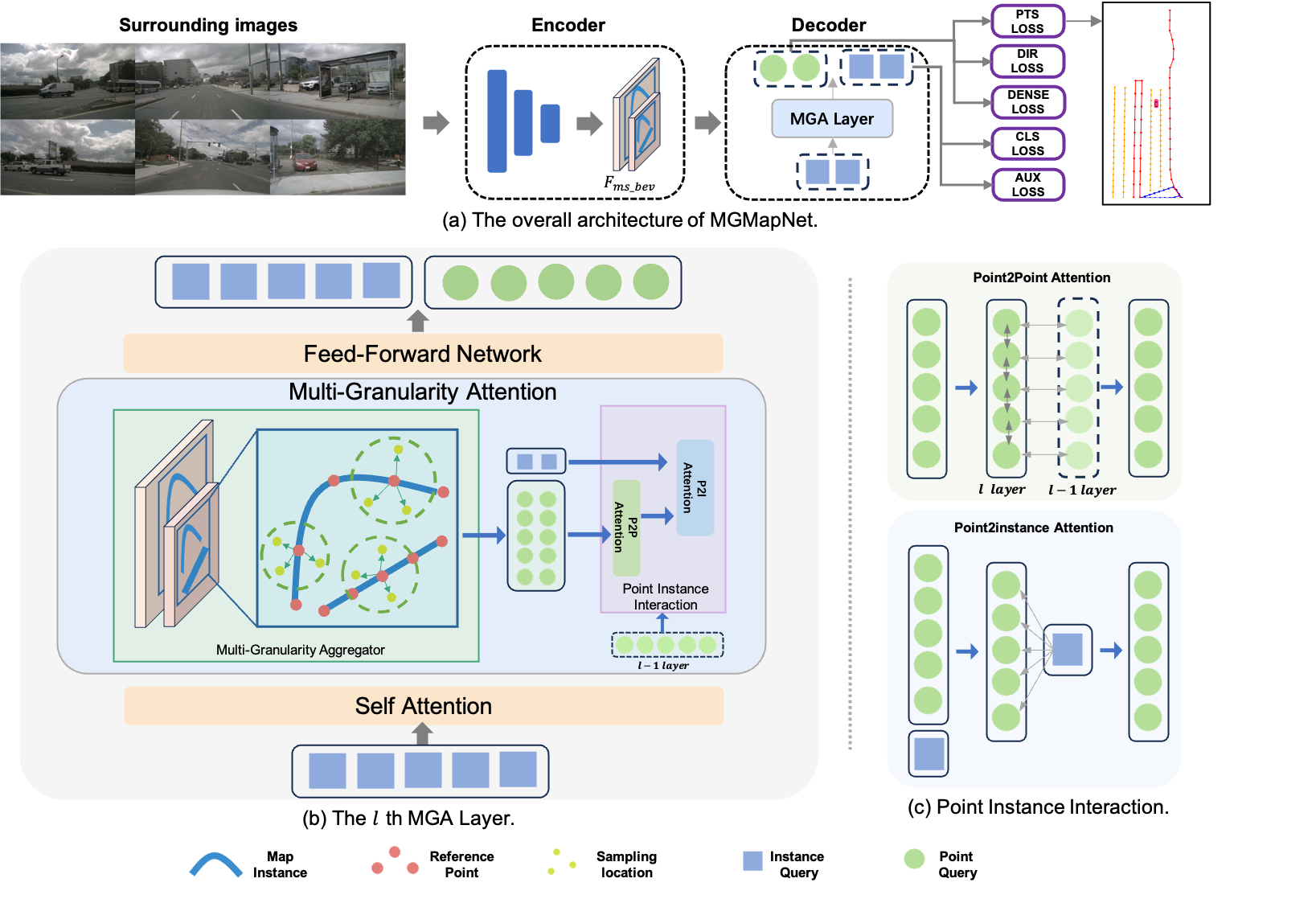}
    \caption{Overview of the MGMapNet. 
    (a) The overall architecture of MGMapNet starts with multi-view image inputs, which are processed through an encoder and decoder to output vectorized map representations.
    (b) A schematic diagram of the $
l$ th MGA layer. The figure depicts the interactions within the Multi-Granularity Attention, including Multi-Granularity Aggregator and Point Instance Interaction.
(c) Implementation details of Point Instance Interaction, consisting of P2P attention and P2I attention.
} 
    \label{fig:MGMapNet}
\end{figure}

\subsection{Overall Architecture}
The overall network architecture of MGMapNet is depicted in Fig.~\ref{fig:MGMapNet} (a). Similar to other DETR-like end-to-end HD map construction models, MGMapNet comprises a BEV Feature Encoder, responsible for extracting multi-scale BEV features from perspective view images, and a Transformer Decoder, which stacks multiple layers of Multi-Granularity Attention to generate predictions for map elements. The prediction from each layer encapsulates both category and geometry information within the perception range.

\textbf{BEV Feature Encoder} The model takes surrounding-view RGB images as inputs, expressing them as unified perceptual BEV feature representation for subsequent transformer decoder. The unified BEV feature is denoted as $\tens{F}_{bev} \in \mathbb{R}^{C \times H \times W}$ , where $C, H, W$ represent the feature channels, height, and width of the BEV feature, respectively. 
Given the diverse lengths of map elements, relying solely on a single-scale BEV feature fails to meet the requirements for detecting all elements with different lengths.
Therefore, we employ downsample modules to reduce the spatial resolution of BEV features $\tens{F}_{bev}$ by half, generating $\tens{F}^{'}_{bev} \in \mathbb{R}^{C \times \frac{H}{2} \times \frac{W}{2}}$. More scales might be benificial, but we found two scales are already good enough.  
Both $\tens{F}_{bev}$ and $\tens{F}^{'}_{bev}$ are utilized in the decoder afterwards.
$\tens{F}_{ms\_bev}\in\mathbb{R}^{C\times(\frac{H}{2} \times \frac{W}{2}+H\times W)}$ represents multi-scale BEV features, which are obtained by concatenating the flattened tensors of $\tens{F}_{bev}$ and $\tens{F}_{bev}^{'}$.
As will be shown in the experiments in the following section, the multi-scale BEV features greatly improve the overall performance.

\textbf{Decoder}
The decoder has $L$ layers. Each layer is composed by a Self-Attention, a Multi-Granularity Attention and a Feed-Forward Network as shown in Fig.~\ref{fig:MGMapNet} (b). Multi-Granularity Attention consists of two components: Multi-Granularity Aggregator and Point Instance Interaction. The instance-level query is initialized by learnable parameters which are updated by querying on BEV features, and the point query is generated dynamically by aggregating BEV features.  
After that, a Point Instance Interaction is employed to carry out the mutual interaction between local geometry information and the global category information.
The details of Multi-Granularity Attention is described in the following section.

\subsection{Multi-Granularity Attention}

As Fig.~\ref{fig:MGMapNet} (b) shown, Multi-Granularity Attention comprises two principal components: Multi-Granularity Aggregator and Point Instance Interaction. 

\subsubsection{Multi-Granularity Aggregator}
\label{sec:Multi-Granularity}

In Multi-Granularity Aggregator, instance-level queries interact with the multi-scale BEV features and point-level queries are generated. Specifically, we improve Mutli-Head Deformable Attention~\cite{zhu2020deformable} with multiple reference points for each query to aggregate long-range features from multi-scale BEV features. To improve readability, we omit the subscript $m$ for the index of multiple heads $M$ in the operator of Multi-Granularity Aggregator.

More specifically, the Multi-Granularity Aggregator takes as input the instance-level queries $\tens{Q}_{ins} \in \mathbb{R}^{N_q \times C}$ in the first layer, along with the point-level queries $\tens{Q}_{pts} \in \mathbb{R}^{N_q \times N_p \times C}$ and the reference points $\tens{RF} \in \mathbb{R}^{N_q \times N_p \times 2}$ in the subsequent layers.
\( N_q \) is the total number of instance-level queries, and \( N_p \) is the total number of points belonging to an instance.
Noted that the reference points in the first layer are predicted by $\tens{Q}_{ins}$ and the reference points in subsequent layers are updated by reference points from the previous layer in the form of Eq.~\ref{equ:ref}.
\begin{equation}
\label{equ:ref}
\left\{
    \begin{array}{l}
	\displaystyle \tR\tF^{l} = \operatorname{MLP}(\tens{Q}_{ins}^{l}),l=0\\
	\displaystyle \tR\tF^{l} = \operatorname{sigmoid}(\operatorname{sigmoid}^{-1}(\tens{RF}^{l-1})+\operatorname{MLP}(\tens{Q}_{pts}^{l})),l>=1\\
    \end{array}
    \right.
\end{equation}
where $l$ represents the current $l$-th layer, $\operatorname{sigmoid}$, $\operatorname{sigmoid}^{-1}$ refers to the sigmoid and inverse sigmoid activation function and $\operatorname{MLP}$ stands for Multi-Layer Perceptron layer.

Since an instance is represented as a point sequence, the position encoding is added to the instance-level query. Given the location of reference point $\tens{RF}$, we employ $\tens{RF}$ to generate positional encoding $\tens{PE}_{ref}$:
\begin{equation}
\begin{aligned}
\tens{PE}_{ref}^{l-1}=\operatorname{MLP}_{ref}^{l-1}(\tens{RF}^{l-1}),
\end{aligned}
\end{equation}
where $\operatorname{MLP}_{ref}^{l-1}$ is a projection layer used to generate the positional embedding from reference points.

We allocate $N_{rep}$ sampling points to each reference point where the features are aggregated from to enhance the feature for the reference point. The location offset $\Delta\tS$ of sampling points w.r.t the reference point and the associated weights  $\tW$ are computed by combining the instance-level queries $\tQ_{ins}$ and $\tP\tE_{ref}$ as follows:
\begin{equation}
\begin{aligned}
\Delta\tS^{l}&=\operatorname{Sampling\_Offset}(\tQ_{ins}^{l-1}+\tP\tE_{ref}^{l-1}) \in \tR^{N_{q} \times N_{p} \times N_{rep} \times 2} ,\\
\tens{W}^{l}&=\operatorname{Weight\_Embed}(\tens{Q}_{ins}^{l-1}+\tens{PE}_{ref}^{l-1}) \in \mathbb{R}^{N_{q} \times N_{p} \times N_{rep}} ,\\
\tens{S}^{l}&=(\tens{RF}^{l-1}+\Delta\tens{S}^{l})\in\mathbb{R}^{N_{q} \times N_{p} \times N_{rep} \times 2},
\end{aligned}
\end{equation}

where $\tR\tF^{l-1}$ is expanded accordingly to match the shape of $\Delta\tS^{l}$. By utilizing the sampling offset and the reference point, sampling locations $\tS^{l}$ are updated by adding $\tR\tF^{l-1}$ and $\Delta\tS^{l}$.

Subsequently, \( \tQ_{ins} \) and \( \tQ_{pts} \) are generated by the weighted sum of sampled features:
\begin{equation}
\begin{aligned}
\tens{W}_{ins}^{l}&=\mathop{\operatorname{softmax}}\limits_{(j,k)\in(N_{p},N_{rep})}\left(\tW_{j,k}^{l}\right)\in \mathbb{R}^{N_q\times (N_{p}\times N_{rep})},\\
\tens{W}_{pts}^{l}&=\quad\mathop{\operatorname{softmax}}\limits_{k\in N_{rep}}\quad\left(\tW_{j,k}^{l}\right)\in \mathbb{R}^{N_q\times N_{p}\times N_{rep}},\\
\tQ_{ins}^{l}&=\sum_{j=1}^{N_{p}}\sum_{k=1}^{N_{rep}}\left[ \tW_{ins}^{l}  \operatorname{sampling(\tF_{ms\_bev},\tS^{l}_{j,k})}\right]\in\mathbb{R}^{N_q\times C},\\
\tQ_{pts}^{l}&=\sum_{k=1}^{N_{rep}}\left[ \tW_{pts}^{l}  \operatorname{sampling(\tF_{ms\_bev},\tS^{l}_{j,k})}\right]\in\mathbb{R}^{N_q\times N_{p} \times C},
\end{aligned}
\end{equation}
where $j$ is the index for the $N_p$ points on an instance, $k$ is the index among the $N_{rep}$ sampling points assigned to the reference point, $\tens{W}_{ins}^{l}$, $\tens{W}_{pts}^{l}$ denotes the softmax normalized weight across $N_{p}\times N_{rep}$, $N_{rep}$ of $\tens{W}_{j,k}^{l}$, respectively and $\operatorname{sampling}$ denotes the  bi-linear sampling operator. 

Through Multi-Granularity Aggregator, \( \tQ_{ins} \) and \( \tQ_{pts} \) are generated from multi-scale BEV features, capturing both global and local information for each map element. 
Compared with Multi-Point Attention as proposed in StreamMapNet~\cite{streammapnet}, our method incorporates point-level queries directly from the multi-scale BEV features by sampling points, which enhances the accuracy of predicted geometry points. In addition, compared with point-level alone representations such as MapTR~\cite{liao2022maptr} and MapTRv2~\cite{maptrv2}, our model updates instance-level queries with sampled point features, which effectively captures the overall category and shape information of road elements.

\subsubsection{Point Instance Interaction}
The Point Instance Interaction is designed with the intention of enhancing positional and categorical information interaction between two different granularities of queries. 
As illustrated in Fig.~\ref{fig:MGMapNet}(c), Point Instance Interaction comprises two distinct attention operators: P2P (point-to-point) attention and P2I (Point-to-Instance) attention.

Concurrently, the sampling locations $\tens{S}^{l}$ and attention weights $\tens{W}_{ins}^{l}, \tens{W}_{pts}^{l}$ obtained from Multi-Granularity Aggregator in the $l$-th layer are flattened and concatenated to encode positional information in P2P Attention and P2I Attention: 

\begin{equation}
\begin{aligned}
\textbf{PE}^{l}_{ins}&=\operatorname{MLP}_{ins}^{l}(\tens{S}^{l}, \tens{W}_{ins}^{l}), \\
\textbf{PE}^{l}_{pts}&=\operatorname{MLP}_{pts}^{l}(\tens{S}^{l}, \tens{W}_{pts}^{l}), 
\end{aligned}
\end{equation}
where $\operatorname{MLP}_{ins}^{l}$ and $\operatorname{MLP}_{pts}^{l}$ are MLPs for instance-level queries and point-level queries respectively. $\tens{PE}_{ins},\tens{PE}_{pts}$ are the corresponding generated position embedding.

\paragraph{P2P Attention}  
As the coordinates of map elements are refined based on the point-level queries in previous Multi-Granularity Attention layer,
these point-level queries play a pivotal role in predicting coordinates in the current layer. Hence, the P2P Attention module is devised to include point-level queries from both the current $l$-th layer and previous $(l-1)$-th layer as inputs of the attention layer. Formally:

\begin{equation}
    \left\{
    \begin{array}{l}
	\displaystyle \tens{Q}_{pts}^{l'}= \operatorname{SA}(\tens{Q}_{pts}^{l}+\textbf{PE}^{l}_{pts}),l=0\\
	\displaystyle \tens{Q}_{pts}^{l'} = \operatorname{CA}(\tens{Q}_{pts}^{l}+\textbf{PE}^{l}_{pts},\tens{Q}_{pts}^{l-1}+\textbf{PE}^{l-1}_{pts}),l>=1\\
    \end{array}
    \right.
\end{equation}
It is important to note that since the first Multi-Granularity Attention layer does not have previous decoder layer, self-attention operation only conducts in the current point-level queries. And in the P2P Attention of following Multi-Granularity Attention layer, previous point-level queries $\tens{Q}_{pts}^{l-1}$ and current generated point-level queries $\tens{Q}_{pts}^{l}$are mixed before P2P Attention.

\paragraph{P2I Attention}
Following the P2P Attention, P2I Attention operation achieves information interaction among different granularities. Point-level queries exchange geometry information with instance-level queries using cross-attention:

\begin{equation}
\begin{aligned}
\tens{Q}_{pts}^{l''}=\operatorname{CA}(\tens{Q}_{pts}^{l'}+\textbf{PE}_{pts}^{l},\tens{Q}_{ins}^{l}+\textbf{PE}_{ins}^{l}).
\end{aligned}
\end{equation}

Finally, point-level queries belonging to the same instance-level queries are aggregated to update corresponding instance-level queries as follows:

\begin{equation}
\begin{aligned}
                \tens{Q}_{ins}^{l'}=\operatorname{MLP}_{agg}^{l}(\sum_{j=1}^{N_{p}}\tens{Q}_{pts,j}^{l^{''}}), 
\end{aligned}
\end{equation}
where $j$ represents the index of $N_{p}$.

\paragraph{Output} Ultimately, point-granularity queries are utilized for point location prediction using a MLP as the regression head, while instance-granularity queries are employed for predicting the categories of map elements using another MLP.
 In summary, by utilizing the Multi-Granularity Aggregator and Point Instance Interaction, Multi-Granularity queries are generated and updated. Meanwhile, the geometry and category of each map element can be effectively perceived.

\section{Experiments}
\label{others}

\begin{table}[t]
  \caption{Comparison to the state-of-the-art on nuScenes val set.}
  \label{table:nus}
  \centering
  {%
  \begin{tabular}{l|c|llll|cc}
    \toprule

    Methods     & Epoch     & $AP_{ped}$ &$AP_{div}$ &$AP_{bou}$ &mAP  &FPS &Params.\\
    \midrule
    HDMapNet~\cite{HDMapNet} &30 & 14.4  & 21.7 & 33.0  &23.0     &- &-\\
    BeMapNet~\cite{Qiao_2023_CVPR} &30 &62.3 &57.7 &59.4 &59.8 &4.3 &-\\
    PivotNet~\cite{Ding_2023_ICCV} &24 &56.5 &56.2 &60.1 &57.6 &9.2 &-\\
    MapTRv2~\cite{maptrv2}     &24   &59.8 &62.4 &62.4 &61.5
    &14.1 &40.3\\
    MGMap~\cite{liu2024mgmap} &24 &61.8 &65.0 &67.5 &64.8 &12 &55.9\\
    MapQR~\cite{liu2024leveraging} &24 &\textbf{68.0} &63.4 &67.7 &66.4 &11.9 &125.3\\
    \textbf{MGMapNet}  &24 &64.7&\textbf{66.1}&\textbf{69.4}&\textbf{66.8} &11.7 &70.1\\
    \midrule

VectorMapNet~\cite{liu2023vectormapnet} & 110 &42.5 &51.4 &44.1 &46.0  &- &- \\
    MapTRv2~\cite{maptrv2}      &110   &68.1 &68.3 &69.7 &68.7 &14.1 &40.3\\
    MGMap~\cite{liu2024mgmap} &110 &64.4 &67.6 &67.7 &66.5 &12 &55.9\\
    MapQR~\cite{liu2024leveraging} &110 &\textbf{74.4} &70.1 &73.2 &72.6 &11.9 &125.3\\
    \textbf{MGMapNet}  &110 &74.3&\textbf{71.8}&\textbf{74.8}&\textbf{73.6} &11.7 &70.1\\
    \bottomrule
  \end{tabular}
}
\end{table}

\begin{table}[]
\small 
\centering
\caption{Performance comparison on NuScenes with IoU-based AP.}
\label{tab:nusc_seg}
\begin{tabular}{l|ccc|c} 
\toprule
Methods &$AP^{raster}_{ped}$ &$AP^{raster}_{div}$ &$AP^{raster}_{bou}$ &$mAP^{raster}$ \\
\midrule
MapVR ~\cite{zhang2024online} \tiny{[NeurIPS2023]} &46.0 &39.7 &29.9 &38.5\\
MGMap ~\cite{liu2024mgmap} 
 \tiny{[CVPR2024]}&\textbf{54.5} &42.1 &37.4 &44.7\\
MGMapNet(ours) &54.0&\textbf{42.7}&\textbf{44.1}&\textbf{46.9}\\
\bottomrule
\end{tabular}
\end{table}

\subsection{Experimental Settings}

\paragraph{nuScenes Dataset.}
nuScenes \cite{caesar2020nuScenes} is a widely recognized dataset in the field of autonomous driving research, providing 1,000 scenes, each captured over a continuous 20-second interval. Each dataset sample incorporates data from six synchronized RGB cameras and includes detailed pose information. 
The perception ranges extend from $-15.0m$ to $15.0m$ along the X-axis and from $-30.0m$ to $30.0m$ along the Y-axis.
For experimental purposes, the dataset is partitioned into 700 scenes comprising 28,130 samples for training purposes, and 150 scenes containing 6,019 samples designated for validation.

\paragraph{Argoverse2 Dataset.} 
The Argoverse2 dataset~\cite{wilson2023argoverse} contains multimodal data from 1000 sequences, including high-resolution images from seven ring cameras and two stereo cameras, as well as LiDAR point clouds and map-aligned 6-DoF pose data.
All annotations are densely sampled to facilitate the training and evaluation of 3D perception models.
Results are reported on the validation set, with a focus on the same three map categories as identified in the nuScenes dataset.

\paragraph{Evaluation Metric.}
In alignment with MapTR~\cite{liao2022maptr}, we have adopted the widely-accepted metric of mean Average Precision (mAP), predicated on the Chamfer distance, a measure frequently employed in HD map construction task. Evaluation thresholds are set at 0.5m, 1.0m, and 1.5m.

\paragraph{Auxiliary Loss.}
We incorporate point loss $\mathcal{L}_{pts}$, classification loss $\mathcal{L}_{cls}$, edge direction loss $\mathcal{L}_{dir}$ and dense prediction losses $\mathcal{L}_{dense}$ from MapTRv2~\cite{maptrv2}. 
In addition, we introduce new auxiliary losses $\mathcal{L}_{aux}$, comprising instance segmentation loss $\mathcal{L}_{ins\_seg}$ and reference point loss $\mathcal{L}_{ref}$ to further improve performance. 

To increase the precision of the sampling locations within Multi-Granularity Aggregator, we add a reference loss $\mathcal{L}_{ref}$ to supervise the sampling points.
Besides, to improve the spatial details of instance-level queries, we generate instance BEV segmentation gt mask $M_{bev}$ to supervise instance-level segmentation prediction.
Hungarian matching results are utilized to eliminate negative instance-level queries' segmentation prediction. The instance segmentation loss, denoted as 
$\mathcal{L}_{ins\_seg}$, is formulated as an ensemble of the cross-entropy loss and the dice loss.

The final loss is defined as the weighted sum of the above losses:
\begin{equation}
\begin{aligned}
\mathcal{L}=\beta_{1}\mathcal{L}_{pts}+\beta_{2}\mathcal{L}_{cls}+\beta_{3}\mathcal{L}_{dir}+\beta_{4}\mathcal{L}_{dense}+\beta_{5}\mathcal{L}_{aux}
\end{aligned}
\end{equation}

\paragraph{Implementation Details.}

Our model is trained on 8 A100 GPUs with batchsize as 2, utilizing an AdamW optimize~\cite{loshchilov2018fixing} with a learning rate of $4 \times 10^{-4}$. We adopt the ResNet50 ~\cite{he2016deep} as our backbone and employ a LSS transformation ~\cite{philion2020lift} with a single encoder layer for feature extraction. 
We also adopt the one-to-many training strategy, consistent with MapTRv2~\cite{maptrv2}.
 The model trains for 24 epochs on the nuScenes dataset and 6 epochs on Argoverse2 dataset. We conduct a long training schedule (110 epochs) on the nuScenes dataset for a fair comparison with previous methods.
 We set $N_q =100 $,$N_{rep}$ = 8, $N_p = 20$, $\beta_{1}=5$, $\beta_{2}=2$, $\beta_{3}=0.005$, $\beta_{4}=3$, $\beta_{5}=3$ as the hyperparameters for all settings without further tuning.

\subsection{Comparisons with State-of-the-art Methods}

\paragraph{Results on nuScenes.}
The default evaluation metric for Vectorized HD Map construction is the Chamfer Distance Average Precision (AP).
Tab.~\ref{table:nus} presents the results on the nuScenes validation dataset, utilizing multi-view RGB images as input. 
In comparison to the SOTA method MapTRv2, our MGMapNet has reached an mAP of 66.8, exceeding it by 5.3 mAP with a training duration 24 epochs. 
After a prolonged training period of 110 epochs, MGMapNet achieved  73.6 mAP, which is still significantly higher than the 68.7 mAP of MapTRv2 by 4.9 mAP and 72.6 mAP of MapQR by 1.0 mAP respectively.

The latest models also use rasterization results and employ IoU-based Average Precision (AP) to evaluate reconstruction performance.
As shown in Tab.~\ref{tab:nusc_seg}. 
We evaluate MGMapNet, which achieves an mAP of 46.9, surpassing both MapVR and MGMap in terms of IoU-based AP.

Experimental results substantiate that the proposed multi-granularity representation, which models both local point information and global instance information, significantly enhances predictive performance in both rasterization and vectorization evaluation metrics.

\begin{figure}[t]
    \centering
    \includegraphics[width=1\textwidth]{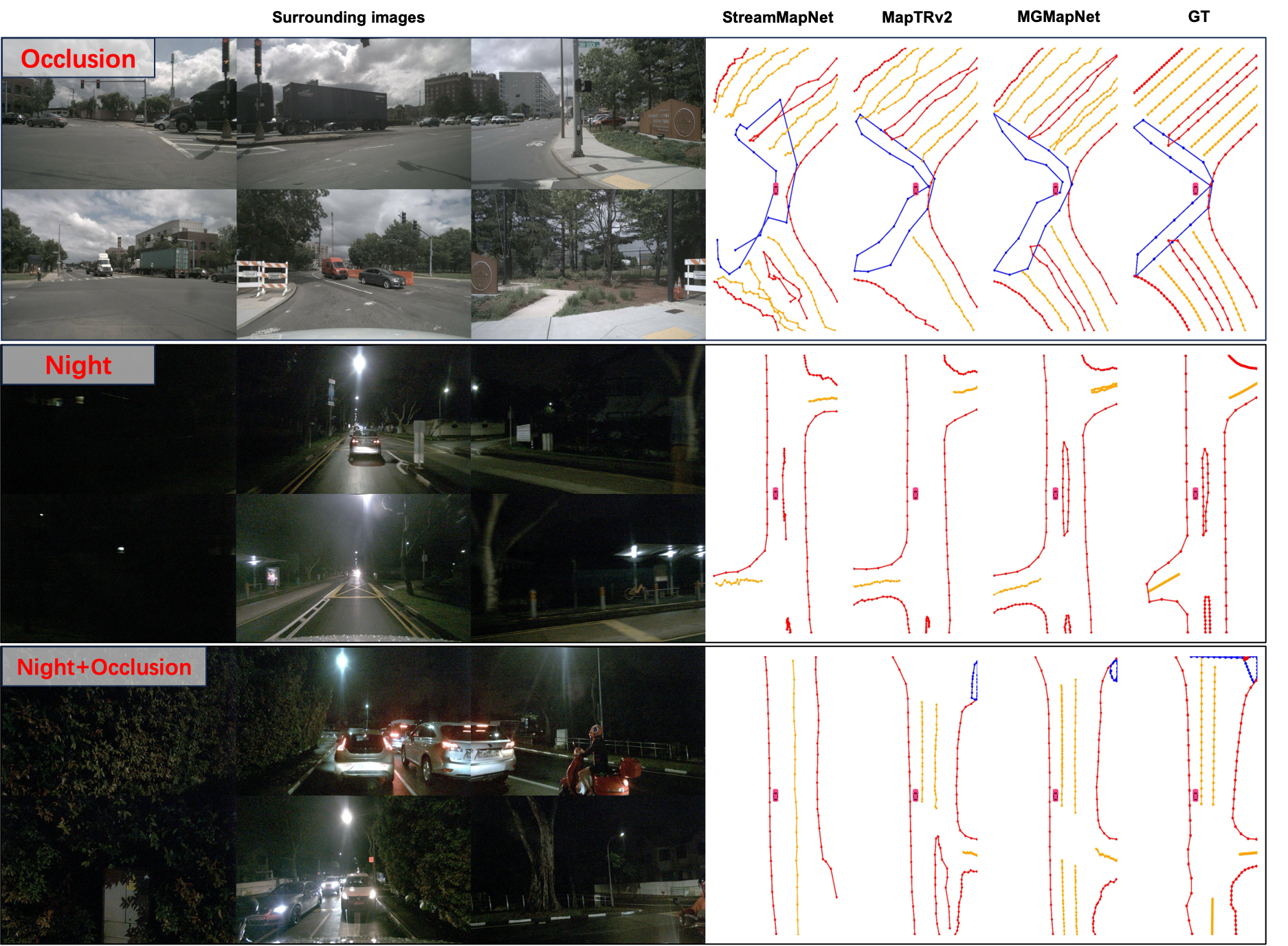}
    \caption{Comparison with SOTAs on qualitative visualization on nuScenes val set.}
    \label{fig:nucs_result}
\end{figure}

Qualitative results are depicted in Fig.~\ref{fig:nucs_result}. 
We select three complex scenarios: daytime vehicles with occlusion, nighttime low-light conditions, and low-light situations with occlusion.
In the first case, MGMapNet exhibits more precise coordinate predictions compared to StreamMapNet and preserves all road elements compared to MapTRv2.
In the second case of nighttime low-light conditions, MapTRv2 struggles to predict the divider on the right side of the vehicle due to its lack of instance-level perception. While StreamMapNet utilizes instance-level queries and identifies the divider, its overall instance positioning accuracy remains inadequate. In contrast, only MGMapNet accurately and completely detects the boundary in these challenging conditions.
The third case of nighttime dense vehicle traffic with occlusion highlights StreamMapNet’s poor detection performance. MapTRv2 encounters two major issues: mislocating the pedestrian path on the right front and misclassifying the rear divider as a boundary, indicating its limitations in instance-level perception.
Conversely, MGMapNet exhibits remarkable robustness, accurately predicting both categories and locations even in low-light conditions and substantial nighttime occlusion.

Qualitative results demonstrate that the proposed MGMapNet effectively mitigates the shortcomings associated with both instance-level and point-level queries, achieving superior accuracy in HD map construction under complex conditions.

\begin{table}[t]
	\centering
	\caption{Comparison to the state-of-the-art on Argoverse2 val set.}
        \label{table:av2}
        {
	\begin{tabular}{l|c|llll}
		\toprule
    Methods     &Map dim.      & $AP_{ped}$ &$AP_{div}$ &$AP_{bou}$ &mAP  \\
    \midrule
		HDMapNet~\cite{HDMapNet}  & \multirow{6}{*}{2}   &13.1 &5.7 &37.6 &18.8  \\ 
		VectorMapNet~\cite{liu2023vectormapnet}  &  &38.3 &36.1 &39.2 &37.9  \\
    
		 MapTRv2~\cite{maptrv2}     &     &62.9 &72.1 &67.1 &67.4  \\
   MapQR~\cite{liu2024leveraging} &  &64.3 &72.3 &68.1 &68.2 \\
 HIMap~\cite{zhou2024himap} &   &\textbf{69.0} &69.5 &70.3 &69.6\\
   \textbf{MGMapNet}  &  &67.1&\textbf{74.6}&\textbf{71.7}&\textbf{71.2}\\
        \midrule
        VectorMapNet~\cite{liu2023vectormapnet} &\multirow{5}{*}{3}  &36.5 &35.0 &36.2 &35.8\\
     MapTRv2~\cite{maptrv2}     &      &60.7&68.9&64.5 &64.7 \\
     MapQR~\cite{liu2024leveraging} &  &60.1 &71.2 &66.2 &65.9 \\
          HIMap~\cite{zhou2024himap} &  &\textbf{66.7} &68.3 &70.3 &68.4\\
    \textbf{MGMapNet}  &  &64.7&\textbf{72.1}&\textbf{70.4}&\textbf{69.1}\\

     \midrule
	\end{tabular}
}
\end{table}

\paragraph{Results on Argoverse2.}
On the more complex Argoverse2 dataset, the performance of MGMapNet remains competitive.
Tab.\ref{table:av2} presents our results on the Argoverse2 validation dataset for 6 epochs. 
The Argoverse2 dataset provides two configurations for the representation of points: 2D and 3D point coordinates.
We conduct experiments on both configurations and achieve mAP scores of 71.2 and 69.1 mAP in 6 epochs, respectively.
This represents an improvement of 3.8 and 4.4 mAP respectively comparing with MapTRv2.
Compared to the latest HIMap, which achieves 69.6 and 68.4 mAP in 2D and 3D configurations respectively, MGMapNet still surpasses it by 1.6 and 0.7 mAP.
The results from other methods are sourced from the original paper, and the experimental results demonstrate the competitiveness of MGMapNet.

\paragraph{Efficiency comparison.} 
We conduct a comprehensive efficiency analysis of several open-source models, focusing primarily on frames per second (FPS) and model parameters to substantiate the efficacy of the models.
As demonstrated in the last two columns of Tab.~\ref{table:nus}, our model achieves an FPS of 11.7, which is comparable to the latest models, MapQR and MGMap. 
It is slightly lower than MapTRv2 while outperforming methods like PivotNet.
The model parameters are 70.1 MB, which is lower than MapQR’s 120.3 MB but slightly higher than MGMap’s 55.9 MB.

\subsection{Ablation Study}
We conduct ablation experiments on the nuScenes validation dataset under the 24 epoch training setting, examining the effectiveness of the Multi-Granularity Attention, as well as the incremental impact of the strategy optimization on the model's performance.
The influence of each component in MGMapNet is demonstrated in Tab.~\ref{table:ablation_MGA}.

\begin{table}
\caption{Effectiveness of key designs in Multi-Granularity Attention. 
}

\label{table:ablation_MGA}
  \centering
  {
  \begin{tabular}{c|c|c|l}
    \toprule
    \multirow{2}{*}{Method}     &\multicolumn{2}{|c|}{Point Instance Interaction}   & \multirow{2}{*}{mAP} \\ \cline{2-3}
     &P2P attention &P2I attention\\
    \midrule
    Multi-Point Attention&-&-&59.6\\
    \midrule
   \multirow{4}{*}{Multi-Granularity Attention}     &-&-&62.7 \\
   & \checkmark  & -    &64.8  \\
      & -  &\checkmark  & 65.0  \\
    &\checkmark &\checkmark &66.8\\
    \bottomrule
  \end{tabular}
  }
\end{table}

\begin{table}
  \caption{Ablation study of each optimization strategy.}
  \label{table:ablation}
  \centering
  {
  \begin{tabular}{l|l|l}
    \toprule
    Exp.     & Method     & mAP \\
    
    \midrule
     & Multi-Point Attention &55.9\\
     \midrule
    (a)     & Multi-Granularity Attention &63.6 (+7.7)\\
    
    (b)     & \quad + Aux. Loss       &64.4 (+0.8)  \\
    (c)     & \quad + Multi-Scale      &65.0 (+0.6)  \\
    (d)     & \quad + Reference Point PE       &66.2 (+1.2)  \\
    (e)     & \quad + Add Query Number    & 66.8 (+0.6)  \\    
    \bottomrule
  \end{tabular}
}
\end{table}

\paragraph{Multi-Granularity Attention.}

Tab.\ref{table:ablation_MGA} illustrates the comparison between MPA and MGA, as well as Point Instance Interaction.
We initially employed MPA as the fundamental module with strategy optimizations, achieving an mAP of 59.6.
By replacing MPA with MGA and introducing more appropriate queries, it captures fine-grained point and coarse-grained instance features. This enhancement facilitates a more nuanced and precise perception, ultimately achieving 66.8 mAP and leading to \textbf{7.2} improvement in mAP.
Additionally, only using the Multi-Granularity Aggregator, the mAP is 62.7, indicating that the multi-granularity representation has led to an mAP increase of 3.1 compared to 59.6 for MPA.
Further, when the P2P and P2I attention are introduced in the Point Instance Interaction, the mAP increased by 2.1 and 2.3 respectively, reaching 64.8 and 65.0. The simultaneous application of these improvements has boosted the model’s performance to 66.8 mAP, an increase of \textbf{4.1} mAP. 
This highlights the significance of both attention modules in enhancing the intrinsic relationships between the two granularities and improving model performance.

\paragraph{Strategy Optimization.}

As shown in Tab.~\ref{table:ablation}, we also investigate the effectiveness of other strategies used in MGMapNet. 
Experiment (a) demonstrates that Multi-Granularity Attention, as a replacement for Multi-Point Attention, achieves a mAP of 63.6, resulting in a 7.7 mAP increase compared to the 55.9 mAP of Multi-Point Attention.
Meanwhile, Experiment (b) reveals that the inclusion of the auxiliary loss results in an improvement of the mAP by 0.8.
Experiments (c), (d), and (e) illustrate the effectiveness of using multi-scale approaches, adding the reference point positional encoding, and increasing the number of queries, which yield gains of 0.6, 1.2, and 0.6 mAP, respectively. By optimizing with these strategies, our MGMapNet achieves 66.8 mAP, representing state-of-the-art performance.

\section{Conclusion and Discussion}

In this paper, multi-granularity representation is proposed, enabling the end-to-end vectorized
HD Map construction using coarse-grained instance-level and fine-grained point-level queries. 

Through the designed Multi-Granularity Attention, category and geometry information is exchanged.
Our proposed MGMapNet has achieved state-of-the-art (SOTA) single-frame performance on both the nuScenes and Argoverse2 datasets.

However, our primary focus is on improving the quality of HD Map Construction. 
Addressing real-time performance is a promising direction for future optimization.
In addition, exploring some temporal approaches as priors is also a direction worth considering.
The mechanism of Multi-Granularity Attention is generic, and it is worth trying to determine its effectiveness in topological prediction or other autonomous driving tasks.

\bibliographystyle{unsrt}  

\newpage

\bibliography{references}

\newpage
\appendix
\section{Appendix}

In this appendix material, we provide additional analysis of the proposed MGMapNet, including:
\begin{itemize}
\item Visualization.
\item Hyperparameter Experimentation.
\item Multi-Scale BEV feature Performance.
\item Ablation study on the hyperparameters Loss Function $\mathcal{L}_{aux}$.
\item Long-term training results of Argoverse2.
\end{itemize}

\subsection{Supplementary experiments.}

\begin{figure}[h]
    \centering
    \includegraphics[width=1\textwidth]{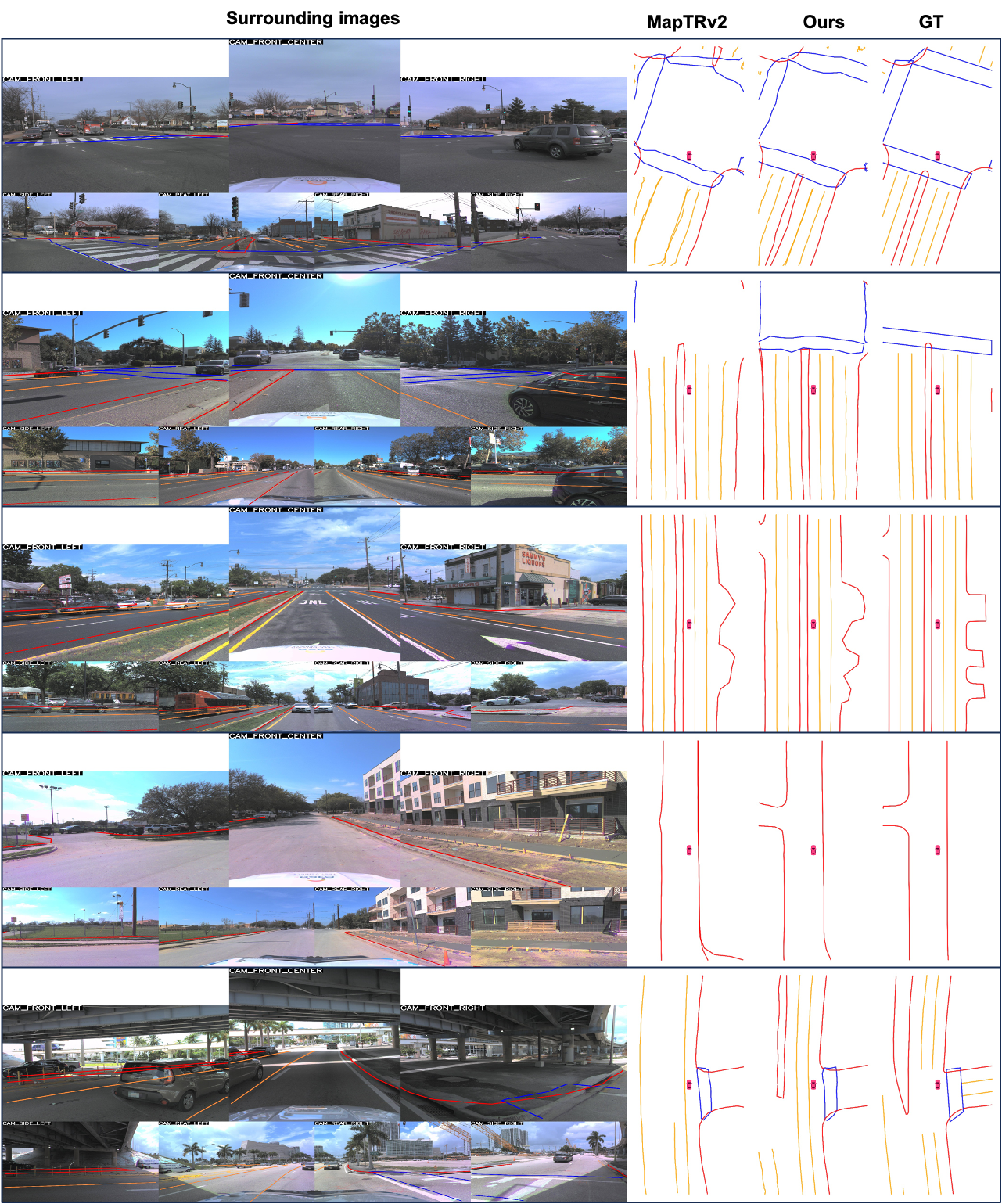}
    \caption{Visualization on Argoverse2 val set.}
    \label{fig:supp_av2}
\end{figure}

\paragraph{Visualization.} We have visualized some comparative results in the supplementary material, as shown in Fig.~\ref{fig:supp_nusc} for the nuScenes dataset and Fig.~\ref{fig:supp_av2} for the Argoverse2 dataset.

In the visualization for the nuScenes dataset, from left to right, each column corresponds to the surround view images, StreamMapNet, MapTRv2, and the proposed MGMapNet, respectively.
From top to bottom, each row represents a sample.
It can be observed that in the initial three samples, MapTRv2 was unable to detect all the pedestrian crossings completely. In the fourth sample, the structure of the pedestrian crossing is inaccurate.
Meanwhile, the results from StreamMapNet indicate that despite the detection of the majority of map instances, the instability of the shapes compromises their ability to accurately detect map elements.

Similar phenomena can also be observed in the visualization for the Argoverse2 dataset.
We can see that in the first and second examples, MapTRv2 missed the middle boundary of the road and the pedestrian crossing in front of the vehicle respectively, while MGMapNet successfully detected both. In the third and fourth samples, MapTRv2 did not correctly detect the exit on the left side and instead interpreted it as a continuous boundary. In the fifth example, in poor lighting conditions, MGMapNet successfully detects the boundary on the left.

In summary, our MGMapNet demonstrates superior performance compared to MapTRv2 on the quality of HD Map Construction.


\paragraph{Hyperparameter Experimentation.}
We conducted experiments on the hyperparameters within the model, including those for $N_q$ and $N_{rep}$, as shown in Tab.~\ref{table:supp_N_rep} and Tab.~\ref{table:supp_N_q}. The parameter $N_q=100$ and $N_{rep}=8$ that we selected represents the optimal configuration.
We believe that too few points are insufficient to describe local detail, while too many points can increase learning difficulty and reduce performance.
\begin{table}[h]
  \caption{Influnence of repeat number $N_{rep}$, the $N_{rep}$ is set as 8.}
  \label{table:supp_N_rep}
  \centering
  \begin{tabular}{l|lll|l}
    \toprule
    Number    & $AP_{ped}$ &$AP_{div}$ &$AP_{bou}$ &mAP  \\
    \midrule

4  &61.5 &62.3 &67.3 &63.7    \\
8  &64.7 &66.1 &69.4&66.8\\
12 &62.8 &63.6 &67.9 &64.8  \\
    \bottomrule
  \end{tabular}
\end{table}

\begin{table}[h]
  \caption{Influnence of query number $N_{q}$, the $N_{q}$ is set as 100.}
  \label{table:supp_N_q}
  \centering
  \begin{tabular}{l|lll|l}
    \toprule
    Number    & $AP_{ped}$ &$AP_{div}$ &$AP_{bou}$ &mAP  \\
    \midrule

50         &64.9 &66.1 &67.3 &66.2  \\
75 &64.6 &65.9 &69.1 &66.5\\
100 &64.7 &66.1 &69.4 &66.8\\
125 &66.8&63.0&69.6&66.4\\
    \bottomrule
  \end{tabular}
\end{table}

\begin{table}[h]
  \caption{Influnence of Multi-Scale BEV feature.}
  \label{table:supp_MS}
  \centering
  \begin{tabular}{l|lll|l}
    \toprule
    Shape    & $AP_{ped}$ &$AP_{div} $&$AP_{bou}$ &mAP  \\
    \midrule

(200,100)         &64.1 &64.9 &68.7 &65.9  \\
(100,50) &64.6 &65.3 &69.3 &66.3\\
Multi-Scale &64.7 &66.1 &69.4 &66.8\\
    \bottomrule
  \end{tabular}
\end{table}


 

\paragraph{Multi-Scale BEV feature Performance.}
In Tab.~\ref{table:supp_MS}, we conducted ablation experiments on the scale of BEV features in MGMapNet, including two different BEV sizes and multi-scale. The results showed that when the BEV size is 
$200\times 100$ and $100\times 50$, the mAP values are 65.9 and 66.3, respectively. By using multi-scale BEV features, we can capture diverse lengths of map elements, and the mAP reached 66.8.

\paragraph{Ablation study on the hyperparameters Loss Function $\mathcal{L}_{aux}$.}
The configuration of other losses follows the original Maptrv2 scheme. As shown in Tab.~\ref{tab:supp_loss}, when the loss weight is 0, the mAP is 65.8; however, when the weight increases to 3, the mAP increases to 66.8, which is the optimal result, proving the effectiveness of the auxiliary loss. Additionally, the experiments demonstrate that a configuration with a weight of 3 is optimal.

\begin{table}[h]
\small 
\centering
\caption{Ablation Experiments on Loss Function $\mathcal{L}_{aux}$.}
\label{tab:supp_loss}
\begin{tabular}{cccccc} 
\toprule
loss weight &$AP_{ped}$ &$AP_{div}$ &$AP_{bou}$ &mAP \\
\midrule
0 &63.7 &65.0 &68.7 &65.8\\
1 &64.0&64.8&69.5&66.1\\
2 &64.0&66.3&68.9&66.4\\
3 &64.7 &66.1 &69.4 &66.8\\
4 &64.5 &65.9 &69.1 &66.5\\
\bottomrule
\end{tabular}
\end{table}

\paragraph{Long-term training results of Argoverse2.}
Most models only report experimental results for 6 epochs, while we present the long-term training results for 24 epochs 2D point coordinates here. As shown in the Tab.~\ref{tab:long_av2}, MGMapNet still performs exceptionally well.

\begin{table}[h]
\small 
\centering
\caption{Long-term training results of Argoverse2.}
\label{tab:long_av2}
\begin{tabular}{cccccc} 
\toprule
Methods &$AP_{ped}$ &$AP_{div}$ &$AP_{bou}$ &mAP \\
\midrule
MapTRv2~\cite{maptrv2} &68.3&74.1&69.2&70.5\\
HIMap~\cite{zhou2024himap}&72.4 &72.4 &73.2 &72.7\\
MGMapNet &71.3 &76.0 &73.1 &73.6\\
\bottomrule
\end{tabular}
\end{table}

\begin{figure}[h]
    \centering
    \includegraphics[width=1.0\textwidth]{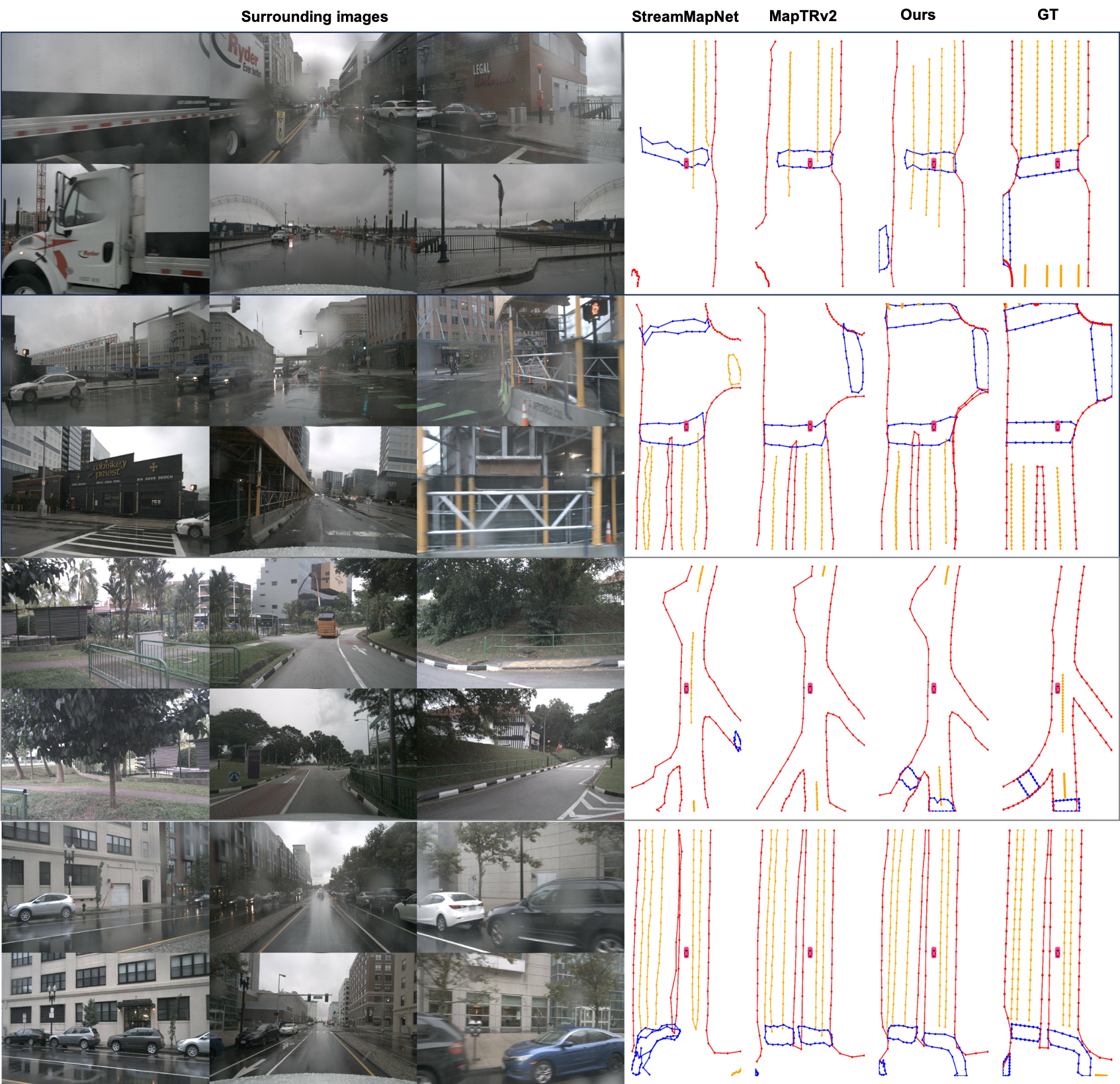}
    \caption{Visualization on nuScenes val set.}
    \label{fig:supp_nusc}
\end{figure}

\end{document}